# Navigating Assistance System for Quadcopter with Deep Reinforcement Learning

Tung-Cheng Wu
Department of Enginnering Science
National Cheng Kung University
Tainan, TAIWAN
n96061070@mail.ncku.edu.tw

Shau-Yin Tseng
Information and Communications
Research Laboratories
Industrial Technology Research Institute
Hsinchu, TAIWAN
tseng@itri.org.tw

Chin-Feng Lai
Department of Enginnering Science
National Cheng Kung University
Tainan, TAIWAN
cinfon@ieee.org

Chia-Yu Ho
Department of Enginnering Science
National Cheng Kung University
Tainan, TAIWAN
n96061119@mail.ncku.edu.tw

Ying-Hsun Lai
Department of Computer Science and
Information Engineering
National Taitung University
Taitung, TAIWAN
yhlai@nttu.edu.tw

*Abstract*—In this paper, we present a deep reinforcement learning method for quadcopter bypassing the obstacle on the flying path. In the past study, algorithm only control the forward direction about quadcopter. In this letter, we use two function to control quadcopter. One is quadcopter navigating function. It is based on calculating coordination point and find the straight path to goal. The other function is collision avoidance function. It is implemented by deep Q-network model. Both two function will output rotating degree, agent will combine both output and turn direct. Besides, deep Q-network can also make quadcopter fly up and down to bypass the obstacle and arrive at goal. Our experimental result shows that collision rate is 14% after 500 flights. Based on this work, we will train more complex sense and transfer model to real quadcopter.

Keywords — deep reinforcement learning, obstacle avoid, quadcopter control

## I. Introduction

After DeepMind published their result about using deep reinforcement learning to play Atari [1], causing a large influence for reinforcement learning. In traditional reinforcement learning, huge input will cause model too hard to calculate Q-table and converge to best action. In deep reinforcement learning, replacing Q-table with neural network to calculate Q-value. Researcher avoids creating a very big Q-table and get the Q-value from the neural network. This change make deep Q learning can handle large input (for example, image data).

Deep reinforcement learning has been deployed to many different situation. For instance, DeepMind has a great success on Go [2]. In addition, there are also play well in other game, like first person shooter (FPS) game[3]. This research point out that their agent can play FPS better than human no matter in single player or multiple players.

Besides playing game, some study use deep reinforcement learning to control quadcopter. There are two controlling issue. One is making quadcopter stable and the other is collision avoidance. In quadcopter stable, researchers' purpose is making quadcopter return to stable state, when encountering strong wind, collision and so on. The most recent study in quadcopter stable presented a method [4] which is implemented by policy network and value network[1].

In collision avoidance, researchers aim to avoid quadcopter colliding intensively, or avoid potential collision early. It was hot issue in control robot and autopilot [5]–[7]. Nonetheless, robot and car can only move in horizontal, but quadcopter can fly up and down. In some situation (like wall), vehicle can't pass barrier, but quadcopter can. Some studies use reinforcement learning to implement quadcopter collision avoidance[8], in spite of the fact that they only present a method to reduce the velocity when quadcopter collided obstacle.

In this paper, we present a quadcopter agent for bypassing obstacles in 3D environment. This agent has two parts. One is quadcopter navigation function. This function will find shortest straight path from quadcopter to goal. The other function is collision avoidance function. This function is a deep Q-network, and it will plan the route to bypass obstacle.

Our 3D environment is based on unreal engine with AirSim [9] which is a module presented by Microsoft. With this module, we can easily simulate quadcopter flies in city.

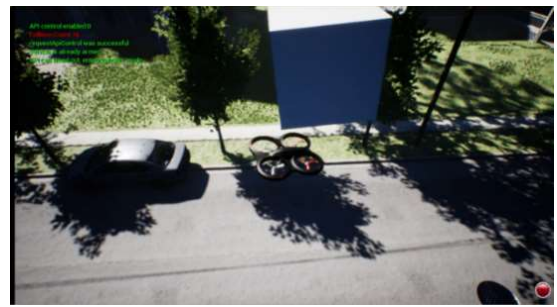

Fig. 1. A screenshot of AirSim
(Quadcopter was bypassing the obstacle (a white cube))

## II. Background

We will give a brief introduction about deep Q-network model and AirSim.in following section.

---
[1] In reinforcement learning, there are two main ways to differentiate agent. First way is rewarded object, there are value method (reward value) and policy method (reward policy). Second way is whether agent with model or not., there are model-free and model-based. In that paper, researcher presented both model-based policy method and model-based value method in their letter.

## A. Deep Q-Network

Reinforcement learning is a learning policy for an agent interacting with unknown environment. Comparing to deep learning, reinforcement learning doesn't have big data to train model. This model usually learns from environment step by step. At each step, agent observes environment and gets current state $S_t$, agent will decide the action $a_t$ with policy $\pi$, and acquire reward $r_t$ from reward function. The final purpose of agent is to maximize the discounted reward $R_t$.

$$R_t = \sum_{t'=t}^{T} \gamma^{t'-t} r_{t'} \quad (1)$$

where $T$ represents the terminal step, $\gamma$ is discounted rate that determines how importance of the future rewards. In Q-learning, we will calculate the Q-value for each action in this state with the policy $\pi$ and save the Q-value in Q-table for querying.

$$Q^\pi(S, a) = \mathbb{E}\left[R_t \mid S_t = s, a_t = a\right]$$

In common situation, we choose the action that has the highest Q-value, and defines $Q^*(S, a)$ as the maximum expected return achievable by following any strategy.

$$Q^*(S, a) = \max_\pi \mathbb{E}[R_t | S_t = s, a_t = a, \pi] \quad (2)$$

$$= \max_\pi Q^\pi(S, a) \quad (3)$$

However, choosing the highest Q-value may cause agent can't get new information about environment. $\epsilon$-greedy strategy can deal with this problem. When agent takes action, there is a probability $\epsilon$ to take action randomly instead of doing the action with maximal Q-value. $\epsilon$ isn't a stable value, it will gradually decrease as step increases.

After acting, agent will acquire the reward about action. With reward, agent can update the Q-value saved in Q-table, and making better action next time.

$$Q^\pi(S, a) = (1 - \alpha) \cdot Q^\pi(S_t, a_t) + \alpha \cdot Q^*(S, a) \quad (4)$$

where $\alpha$ is learning rate. Deep Q-Network replaces Q-table to neural network with weights $\theta$, so we also have to update the weights parameters $\theta$. In the other words, we need to adjust parameters to make $Q^\theta$ close to $Q^*$.

Q-network can be trained by loss function $L_i(\theta_i)$

$$L_i(\theta_i) = \mathbb{E}_{S_t, a, R, S_{t+1}}\left[\left(y_i - Q_\theta(s, a)\right)^2\right] \quad (5)$$

where i is current step, and $y_i = R + \gamma \cdot \max_{a_{t+1}} Q_\theta(S_{t+1}, a_{t+1}))$ if i is not the final step. Otherwise, $y_i = R$. Then, we can update with the following gradient:

$$\nabla_{\theta_i} L_i(\theta_i) = \mathbb{E}_{S_t, a, R, S_{t+1}}\left[\left(y_t - Q_\theta(S, a)\right)\nabla_{\theta_i} Q_{\theta_i}(S, a)\right] \quad (6)$$

The above formula could be approximated as:

$$\nabla_{\theta_i} L_i(\theta_i) \approx \left(y_t - Q_\theta(S, a)\right)\nabla_{\theta_i} Q_{\theta_i}(S, a) \quad (7)$$

In general, Q-learning doesn't update in no time. It updates by replay experience[10]. Agent records experience (including state, action, reward, next state). After agent performs enough steps (it may different in different task), agent will randomly select mini-batch from experience. Training model with mini-batch, updating weights. This method called replay experience.

## B. AirSim

Reinforcement learning is a method learning from error. When we train the algorithm, quadcopters may break before we finish training. Thus, we should deploy our navigate assistance system on real quadcopter, after we train our algorithm well in the simulating environment first.

AirSim is an Unreal engine plugin for simulating quadcopter fly in real world. Microsoft design this plugin for researchers trying their algorithm about controlling quadcopters or cars in Unreal. It provides users with several application program interface (API) about moving the quadcopter, getting quadcopter information and so on. API is callable with C++ or python.

Besides, researcher can create their environment or choose an appropriate scenes in Unreal engine for trying algorithm.

## III. METHOD

We design a system make quadcopter find the shortest path to goal and pass by obstacles on the path. We design two functions implement this system.

### A. Quadcopter navigation function (to find the shortest path)

We define the shortest path as the straight line from quadcopter to goal position. To find the shortest path, we need the position of quadcopter and goal. Then we can calculate turning degree for quadcopter to facing goal. In the simulating experiment, we use AirSim coordination to present position. (In the real world, we may use the GPS information).

### B. Collision avoidance function (to pass by obstacles)

On the path to goal, quadcopter may encounter some obstacles that it has to pass by. For the sake of bypassing the barrier, avoiding collision and re-guide to the shortest path. We used the deep-Q network to deal with this problem.

- **Input layer:** We use depth image as the input of the network. Depth image is acquired from AirSim application program interface. The original size of depth image is 256x144. We use python module OpenCV[11] to resize depth image into 80x80. The input state initialize with first depth image as 80x80x4. It is a queue for save depth image. As get new depth image, new one will enter queue and get rid of the oldest one.

- **Convolutional and pooling layer:** After preprocessing the depth image, we enter those image into convolution layer and max pooling layer. Each convolution layer will send into ReLU, and connect with one pooling layer. There are three convolution and pooling layers in our model structure. All convolution layers set stride = 2 and same padding. In addition, all pooling layers set kernel size = 2 and same padding.

- **Fully connected layer:** This layer flatten the output of convolution and pooling layer and we can acquire 1600 nodes. We enter these flatten nodes into fully connected layer, getting 512 output. Sending these 512 nodes into ReLU, and we enter these nodes and output 13 nodes. These nodes are Q-value for 13 actions.

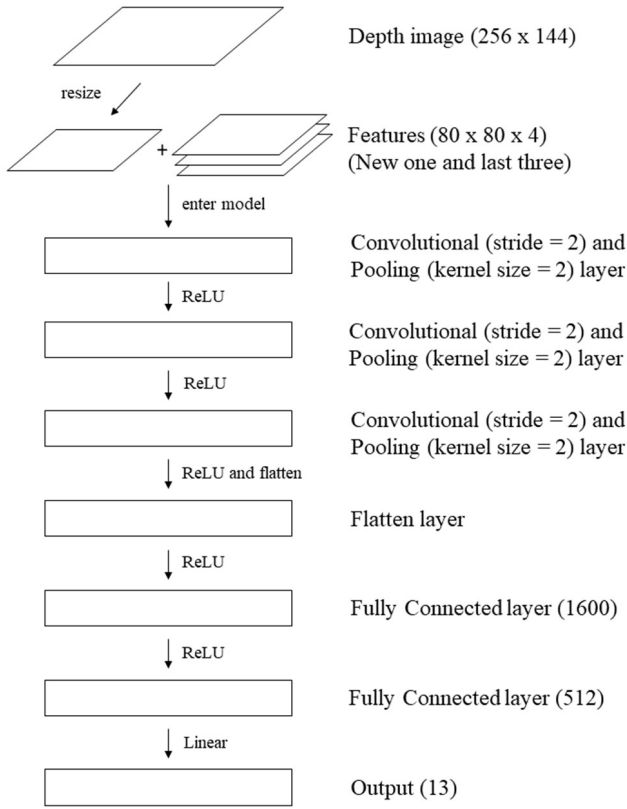

Fig. 2. This plot shows the overall deep Q-leaning model structure in collision avoidance function. The digits in parentheses present number of nodes in that layer.

*C. Action and Exploration Strategy*

As mention in the section II, Q-learning usually chooses the action with $\varepsilon$–greedy strategy. In our method, we output 13 different actions about quadcopter, and set the initial $\varepsilon$ is 0.1. $\varepsilon$ will gradually decrease with increasing of steps. Finally, $\varepsilon$ will decrease to 0.001.

These 13 actions guide quadcopter fly to different direction. Actions can be separated to two category. One move attitude, agent will fly up, fly forward, fly down. The other is turn angle, agent will turn left 15°, turn left 5°, turn right 5°, and turn right 15°. Combining the above two category, we can get 12 action. The last action is go forward.

*D. Decide final action with two functions*

We have two functions (quadcopter navigating function and collision avoidance function) in our model. Both models output how many degrees should quadcopter turn and we add the output degrees directly. Thus, agent controls the turning degree by both quadcopter navigating function and collision avoidance function, and adjusts flying attitude altitude by collision avoidance function.

*E. Reward*

After reaching new position, agent will acquire the new information from environment. With this new information, agent will evaluate the last action, and we call this feedback "reward". In our situation, reward is related to the distance between quadcopter and goal:

$$D = \|P_{move} - P_{goal}\|_2 \qquad (8)$$

where $P_{move}$ is the quadcopter position that has taken action, and $P_{goal}$ is goal position. With this two-positional information, we can define our reward function.

$$R = D_{now} - D_{last} - 50 \cdot Collision \qquad (9)$$

where $D_{now}$ is the distance between goal and quadcopter position after agent took action. $D_{last}$ is the distance between goal and quadcopter positon before agent performs action. *Collision* presents whether quadcopter collides with anything; when quadcopter bump with anything in moving, *Collision* will be 1, or it will be 0.

Thus, reward will be positive when action make quadcopter move closer to goal without any collision. In contrast, reward will be negative if action make quadcopter fly away from goal or quadcopter collide with anything.

*F. Network Training*

In the section II, we introduced the Q-network and brief interpret how Q-network update parameters in the neural network. In practical, there are two Q-network in training. One is $Q_{value}$-network that we use to calculate Q-value, and the other is $Qt_{arget}$-network the we only use to acquire y in the mini-batch.

Every 50 steps, we will enter training state. We will randomly select a mini-batch dataset from experience. The parameters in the $Q_{value}$-network will be updated in every training state. However, the parameters in the $Q_{target}$-network will update every 500 steps, and be updated by copy the parameters in $Q_{value}$-network.

## IV. EXPERIMENT

We setup the training environment in Unreal Engine with AirSim plugin. It can import environment module and freely add objects (cube, column…etc.) in simulating environment. We import the "Neighborhood Pack" module as our training scene. Everyone can buy this module from Unreal Engine or use the binary file provided by Microsoft on AirSim Github.

*A. Experimental procedure*

In our experiment, agent will repeat to observe environment, decide action, get reward and record above information in each step. Besides, model will train deep Q-network each 50 steps, and copy the parameters in $Q_{value}$-network to $Q_{target}$-network every 500 steps. Otherwise, agent will directly enter next step.

When quadcopter successfully arrives at goal or collides with obstacle, it will back to starting position. Every times back to starting position, it will count one flight.

Every 100 flights, we will stop training and enter testing phase. In testing phase, quadcopter fly in the same environment module (neighborhood pack), but at the place that quadcopter has not flown before. We test 1000 flights in each testing phase, and record flight performance.

Flight performance includes rewards, steps and collision times. We use these information to calculate average reward and collision percentage. Average reward present how many reward do each steps acquire in one flight. Collision percentage refers to how many flights that has happened collision in this testing phase.

*B. Experimental condition*

We fly quadcopter on the street in simulating environment with obstacle on the path of start to goal. Initial point is set randomly on (0, *N*(0, 5), -2) in AirSim coordination. *N*(0, 5) is normal distribution with mean=0 and standard deviation=5.

*C. Experimental result*

We record model weights each 100 flights until 500 flights. We take this training weight to fly another place that drone doesn't fly before in the same module. Each weights we test 1000 flights and record whether drone collides with obstacle and how many reward will agent get each step in flight. The following plot shows collision percentage of each weights in 1000 flights.

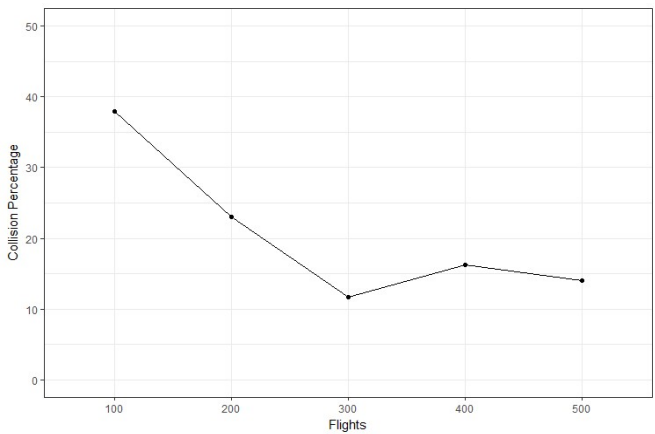

Fig. 3. The collision percentage with training 100 ~ 500 flights. With this plot, we can find that quadcopter gradually find path to bypass obstacle without collision after training 300 flights.

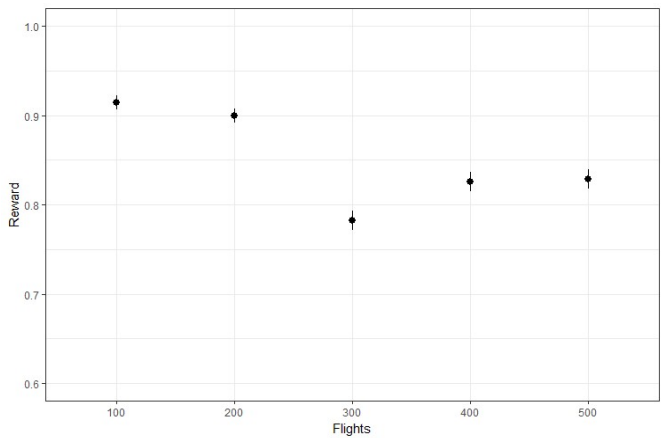

Fig. 4. The average reward with training 100 ~ 500 flights (the line on point is standard error of reward in each flights).

Fig. 4. shows that as training steps go by, agent not only find the path to pass the obstacle but find the path that agent will get the highest reward. Thought 300 flights have the lowest collision percentage, agent gets average stable reward until 400 flights. With the standard error, we can find that agent in 400 and 500 flights have no different in average reward.

## V. CONCLUSION

In our study, We provide a simple model with deep Q-network for quadcopter to find the shortest path to goal with avoiding obstacle. Our model can avoid 86% object obstacle even in the strange path. Compared to before work, our model more simple and make quadcopter to fly in more choices. Based on this work, we hope to train agent in more complex sense to enhance our model and try to transfer training results to real quadcopter in the future.

ACKNOWLEDGE

The authors would like to thank the Ministry of Science and Technology of the Republic of China, Taiwan for supporting this research under Contract 106-2221-E-006 -039.

Besides, this study is conducted under the "Open Service Platform of Hybrid Networks and Intelligent Low-carbon Application Technology Project" of the Institute for Information Industry which is subsidized by the Ministry of Economic Affairs of the Republic of China.